\documentclass[10pt, a4paper]{article}

\usepackage{lrec-coling2024}
\usepackage{amsmath}

\title{Mapping the Past: Geographically Linking an Early 20th Century Swedish Encyclopedia with Wikidata}

\name{Axel Ahlin\textsuperscript{\textdagger}, Alfred Myrne\textsuperscript{\textdagger}\thanks{\textsuperscript{\textdagger} Equal contribution}, Pierre Nugues} 

\address{Lund University \\
         Lund, Sweden \\    
         \{ax5047ah-s, al5247my-s\}@student.lu.se, Pierre.Nugues@cs.lth.se\\\\
         \textit{Paper originally published in the Proceedings of the LREC-COLING, 2024}}

\abstract{
In this paper, we describe the extraction of all the location entries from a prominent Swedish encyclopedia from the early 20th century, the \emph{Nordisk Familjebok} `Nordic Family Book.' We focused on the second edition called \emph{Uggleupplagan}, which comprises 38 volumes and over 182,000 articles. This makes it one of the most extensive Swedish encyclopedias. Using a classifier, we first determined the category of the entries. We found that approximately 22 percent of them were locations. We applied a named entity recognition to these entries and we linked them to Wikidata. Wikidata enabled us to extract their precise geographic locations  resulting in almost 18,000 valid coordinates. We then analyzed the distribution of these locations and the entry selection process. It showed a higher density within Sweden, Germany, and the United Kingdom. The paper sheds light on the selection and representation of geographic information in the \emph{Nordisk Familjebok}, providing insights into historical and societal perspectives. It also paves the way for future investigations into entry selection in different time periods and comparative analyses among various encyclopedias.
 \\ \newline   
}

\begin{document}

\maketitleabstract

\section{Introduction}
The Swedish \textit{Nordisk Familjebok} encyclopedia `Nordic Family Book'  was first published in the late 19th century. It holds a significant place in the history of Swedish literature and is widely regarded as one of the most comprehensive and authoritative encyclopedic works in Swedish. The \textit{Nordisk Familjebok} was originally created with the purpose of providing accurate knowledge and information to the Swedish public.

\subsection{The \textit{Uggleupplagan} Edition}
The inaugural edition, commonly referred to as the ``First Edition,'' was published from 1876 to 1899, spanning 20 volumes. The encyclopedia aimed to encompass a wide array of subjects, including history, geography, science, literature, arts, and more. It featured thoroughly written articles authored by numerous experts in their respective fields. As such, it can be regarded as a text that reflects the dominating worldview in Sweden in the early 20th century.

The second edition of the encyclopedia, also called \textit{Uggleupplagan}, `Owl Edition', was published between 1904 and 1926 and served as a revised and expanded version of the original encyclopedia. This edition comprised 38 volumes and boasted an extensive range of articles, illustrations, maps, and photographs, providing a comprehensive and visually captivating resource. It is the most extensive encyclopedia ever printed in Swedish \citep{projektruneberg}.

The encyclopedia also received a third, fourth, and fifth edition, in the years 1923–1939, 1951-1957, and 1993, respectively. However, the second edition was the most influential, and is therefore the focus of this study. 

In this work, we extracted the named entities from the \textit{Uggleupplagan} encyclopedia and we analyzed the qualitative trends in the selection of entries. More specifically, we focused on the geographic entries.

\subsection{Contributions}
The contributions of this paper are:
\begin{enumerate}
\item We cleaned a 20th century encyclopedia from a raw OCR text and we structured it in the JSON format;
\item We extracted the entries corresponding to named geographic entities using a combination of pre-trained models and fine-tuned classifiers;
\item We mapped these entities to Wikidata items using the entry text and the Swedish description in Wikidata. We first fetched candidates from Wikidata; we ranked them using a semantic cosine similarity; and we selected the candidate with the highest score;
\item We extracted the coordinates from the chosen candidates and we visualized these entries on a map;
\item We published our dataset on GitHub\footnote{\url{https://github.com/axelahlin/uggleupplagan}}.
\end{enumerate}

We found that most geographic entities in the encyclopedia are located in Sweden, as well as in other places in Europe. We also noted a specific focus on Germany and the United Kingdom. This corresponds to countries where Sweden had strong ties at the time of the encyclopedia's production, including historical, scientific, cultural, and economic relations.

\section{Previous Work}
In this work, we used the digital version of an encyclopedia. We applied a binary categorization to extract the location entries, and we then linked these entries to Wikidata. 

\subsection{Digitized Encyclopedias}
A complete digitized version of \textit{Uggleupplagan} is available from Projekt Runeberg\footnote{\url{https://runeberg.org/nf/}}. This project focuses on making older Nordic literature freely available to the public through digital means and \textit{Nordisk Familjebok} is one of them. It is operated as a non-profit organization within the University of Linköping \citep{ne-projektruneberg}.

The digitization had four main steps:
\begin{enumerate}
\item The Runeberg editors first scanned a paper copy of the encyclopedia;
\item They applied an optical character recognition (OCR) to extract the text from the image scans;
\item They structured the text into entries and made them available by headwords. Readers can then access the content from a web browser;
\item Finally, volunteers proofread the machine-generated text, but a majority of entries has not yet been proofread. The scanned pages are openly available for verification.  
\end{enumerate}

\subsection{Entry Categorization}
The \textit{Nordisk Familjebok} entries consist of common as well as proper nouns and the first step is to classify them in these two categories: Whether they are a location or not.

Transformers architectures \citep{Vaswani_Shazeer_Parmar_Uszkoreit_Jones_Gomez_Kaiser_Polosukhin_2017}, and notably the Bidirectional Encoder Representations from Transformers model (BERT) \citep{Devlin_Chang_Lee_Toutanova_2019} have produced the highest performances on classification tasks such as those of the GLUE benchmark \citep{wang-etal-2018-glue}.

The BERT architecture makes it possible to build large pre-trained models that users can then further fine-tune on classification tasks as well as on entity recognition.

\subsubsection{BERT}

BERT utilizes a transformer-based architecture restricted to its encoder part. From an input consisting of a sequence of words, this architecture captures the contextual relationships between them bidirectionally and maps each word to a contextual dense vector. The bidirectional context allows BERT to have a deeper understanding of language and to handle tasks like text classification and named entity recognition more efficiently. 

BERT is pre-trained on a large corpus of English text data and, during this process, it learns to predict masked words within sentences and determines the relationships between sentence pairs.

\subsubsection{KB-BERT} \label{KBBert}
The BERT architecture has been subsequently pre-trained on many languages outside English. \citet{swedish-bert} pre-trained BERT models on Swedish text from the National Library of Sweden that they called KB-BERT. They gathered their corpus from diverse sources, including books, news articles, government publications, Swedish Wikipedia, and internet forums. 

The KB-BERT pre-trained models served as a source of multiple application tasks for Swedish text. \citet{remmer-etal-2021-multi} used them to classify patient records in Swedish while \citet{nik} and \citet{Nyqvist1618067} applied them to named entity recognition. \citet{Bridal1683482} fine-tuned the base KB-BERT model to evaluate the de-identification of Swedish clinical data. Finally, \citet{nielsen2023scandeval} showed that one of the KB-BERT models outperformed all other Swedish models for NER tasks.

\subsection{Entity Linking and Disambiguation}
One task in this work was to link encyclopedia entities to Wikidata items. Wikidata is a multilingual knowledge base associated to Wikipedia. It contains more than 100 million entries at the time of this study and it has become a central repository for authority data disambiguation and linking \citep{tharani2021much}. It contains useful properties for each entry. For instance, entities with geographical locations have the P625 property, which contains the longitude and latitude of the entity. 

Examples of studies using Wikidata include \citet{pratapa2022multilingual}, which linked event descriptions in 44 languages to this repository; \citet{HistoricalNewspapers} has a similar goal with proper nouns in historical documents such as newspapers. 

\subsubsection{Historical Data}
The tasks of classification and entity linking on historical data present unique challenges. This includes OCR errors, different grammar and language rules, and more. \citet{historical} discusses some of these challenges. \citet{nugues2022connecting} linked the named entities of a French dictionary from the turn of the 20th century to Wikidata. The author then visualized the geographic coordinates of the location entities on a map.  

\subsubsection{Named Entity Linking}
Named entity linking often proceeds in three steps \citep{mrini-etal-2022-detection}:
\begin{enumerate}
    \item The first step is to identify the mentions of named entities in text;
    \item For a given mention $M$, extract a list of candidate entities that could match it: $\{E_1, E:2, ..., E_n\}$;
    \item Finally, rank the candidates and select the first one: $\displaystyle{\arg \max_{i} P(E_i|M)}$.
\end{enumerate}

To link entities, \citet{hoffart-etal-2011-robust} used Wikipedia disambiguation links and context similarity scores based on keywords. \citet{francis-landau-etal-2016-capturing} computed a cosine similarity between the embeddings of the mention context and of an entity description taken from a dictionary of entities. \citet{logeswaran-etal-2019-zero} also used an entity mention and an entity description as input. They separated both texts by a \verb=[SEP]= token and they trained a BERT-based classifier to decide if the mention corresponded to the entity. The authors used the classifier score to rank the candidates.

\subsubsection{SBERT}
In our work, the list of candidates consisted of Wikidata items and we ranked the  Wikidata descriptions from the definition of the encyclopedia entry.

We followed \citet{francis-landau-etal-2016-capturing} and we computed an embedding vector for both descriptions, one from Wikidata and the other from the encyclopedia. Nonetheless, instead of convolutional neural networks, we used transformers.

As transformer, we chose Sentence-BERT (SBERT) \citep{Reimers_Gurevych_2019}. SBERT is a modified version of BERT that uses both Siamese and triplet network architectures. SBERT takes a sentence as input and outputs the sentence embedding vector. We can then compare two sentence embeddings using, for example, a cosine similarity. 

\section{Method}
To the best of our knowledge, there exists no  corpus of dictionary entries in Swedish annotated with location and links. We could not then rely on using fully-supervised methods. As we wanted to avoid a completely manual annotation procedure, we designed a semi-automatic pipeline, where we used classification and linking.

\begin{figure*}[tb]
    \begin{center}
    \includegraphics[width=0.9\textwidth]{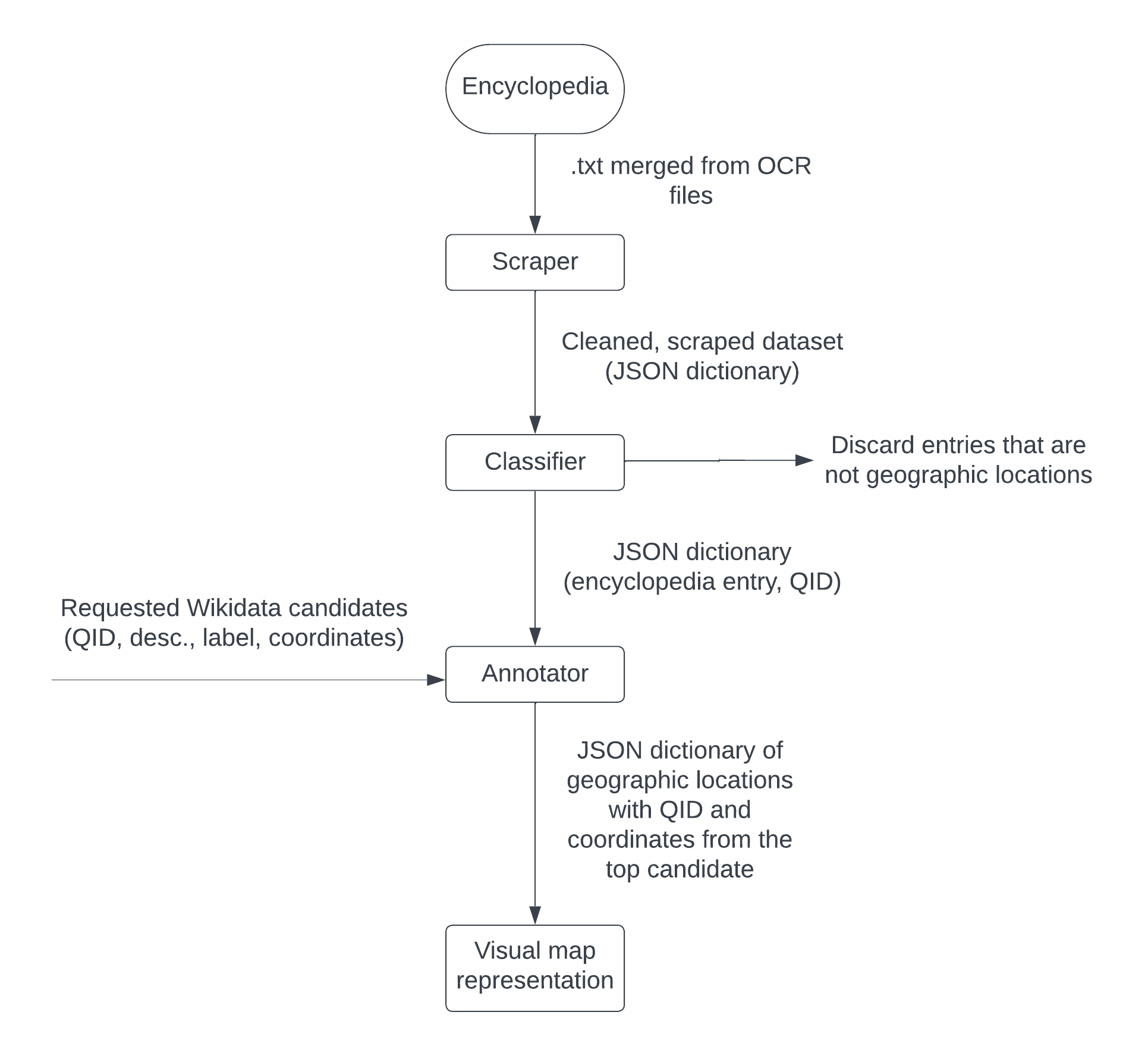}
    \caption{Architecture of the pipeline}
    \label{fig:pipeline}
    \end{center}
\end{figure*}

This pipeline consists of five steps shown in Figure~\ref{fig:pipeline}:
\begin{enumerate}
    \item We first scraped the encyclopedia text from the Runeberg website and we organized the dataset as a JSON file of entries;
    \item We trained a classifier to determine whether an entry is a location or not. As training set, we annotated manually a set of positive and negative samples; we applied this classifier and we discarded the nonlocations;
    \item For each location, we used the encyclopedia text and extracted the headword to query Wikidata for candidate items having the ``geographic location'' property: -- \verb=P625= --;
    \item We computed the cosine similarity between each Wikidata item description in Swedish and the encyclopedia text;
    \item Finally, we extracted and plotted the geographic coordinates of the item with the highest cosine similarity to the encyclopedia text.
\end{enumerate}

The classification step is necessary to the extraction of geographical entries. Instead, one could be tempted to use the headword to fetch all items with the coordinate property \verb=P625=. However, this would not be ideal as it would include a vast number of items that may not be relevant. 

Specifically, some Wikidata items, such as the \emph{Mona Lisa} item, have coordinates assigned for the \verb=P625= property. These coordinates represent the geographic location of the item itself, not necessarily meaning that it actually is a location. Hence, we cannot establish an equivalence between an actual geographic location and a Wikidata item having the \verb=P625= property.

\subsection{Scraping}
We scraped a digitized version of the \emph{Uggleupplagan} from the Projekt Runeberg website\footnote{https://runeberg.org/nf/}. It comprises the 38 volumes available as text files. 
The \emph{Uggleupplagan} contains over 182,000 entries \citep{skandiencyklop}. The entries had an average of 15.51 words, including the entry title word. This resulted in an average length of 102.19 characters per entry. For each entry, we cut the text description after 200 characters and we removed all text after the final period (``.'') character. 

We noticed that longer descriptions displayed a tendency to confuse the classifier by providing superfluous context and leading to incorrect results. This is why we imposed this character limit on the entries as we believe this represents a good trade-off between necessary context for the classifier and classifier performance.

\subsection{Classifier}
For the geographic location classifier, we first annotated a dataset of encyclopedic entries as location or nonlocation with Boolean values. We then fine-tuned the baseline KB-BERT model to categorize the complete set of entries. 

To create the classifier, we applied the pre-trained model to the annotated entries and we used the resulting hidden states to fit a logistic regression model. We then applied the resulting model to categorize all the entries of the encyclopedia. Finally, we purged the entries that were not geographic locations.

\subsection{Entity Candidate Ranking and Linking}
We linked each encyclopedia entry classified as a geographic location to a Wikidata item. A Wikidata item has a unique identifier (QID) consisting of a \verb=Q= prefix and a number, for instance \verb=Q1754= for \textit{Stockholm}. Each Wikidata item has attributes describing it such as a plain text description in multiple languages. These attributes use the \verb=P= prefix and a number, such as \verb=P31= for the \verb=instanceof= property, the type of the item. Wikidata location items have very often geographic coordinates with the \verb=P625= attribute. 

We can search Wikidata with string queries as with a search engine. It then returns a set of items with the Wikidata identifiers (QID).

To link the encyclopedia entries with Wikidata objects, we proceeded in two steps:
\begin{enumerate}
\item For each entry, we extracted the headword or first word and we used it to query Wikidata. We kept up to five items from all the candidates retrieved by the Wikidata query;
\item We then compared the encyclopedia entry text with the Swedish description of each Wikidata item:
\begin{itemize}
\item We encoded the texts, encyclopedia and Wikidata, as embedded vectors with the SBERT model \verb=all-MiniLM-L6-v2=. We believe this model represents a good trade-off between speed and performance. 
\item We computed a cosine similarity for all the pairs.
\end{itemize}
\end{enumerate}

We selected the candidate with the highest cosine similarity and linked the entry to the corresponding Wikidata identifier. 

\subsection{Querying Wikidata}\label{wd}
For all the locations we could extract from the encyclopedia using the classifier, we fetched the corresponding geographic coordinates from Wikidata. 

In addition to the string search queries, we can query Wikidata with the SPARQL graph database language. We used SPARQL to create a query for coordinate retrieval:
\begin{verbatim}
    ?item  wdt:P625  ?coords 
\end{verbatim}
where \verb=?item= is the QID, \verb=wdt:P625=, the coordinate location property, and \verb=?coords=, the coordinates to extract.

Figure~\ref{fig:map} shows the coordinates on a world map of all the location entries in the encyclopedia. 

\section{Results}
\begin{figure}[t]
    \begin{center}
    \includegraphics[width=1.0\columnwidth]{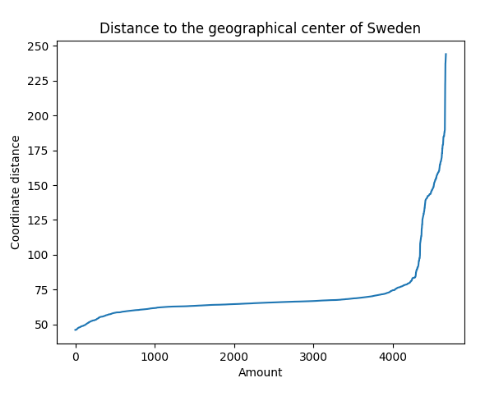}
    \caption{Distribution of entries by geographic distance to Sweden}
    \label{fig:graph}
    \end{center}
\end{figure}

Our scraping resulted in 130,383 entries. 
We applied the classifier to these entries and we obtained 28,284 locations. This means that approximately 21.7 percent of the entries are geographic locations. Once linked, we extracted 17,793 valid coordinates. Figure~\ref{fig:map} shows a geographic representation of the world, where we plotted all extracted coordinate points, while Figure~\ref{fig:graph} shows the coordinate distance from each retrieved coordinate to the geographic center of Sweden.

\section{Evaluation}
\subsection{Classifier}
We evaluated the classifier using a manually annotated test set of 200 previously unseen entries. Table~\ref{fig:conf} shows the confusion matrix and Table \ref{metrics} shows normalized metrics from the statistical report of the classifier. 

\begin{table}[tb]
  \begin{center}
    \begin{tabular}{lrr}
    \multicolumn{3}{c}{Normalized confusion matrix}\\ \\
    \hline
    True label$\backslash$Pred. label&Location&Not location\\
    \hline
    Location&0.93&0.07\\
    Not location&0.06&0.94\\
    \hline
    \end{tabular}
     \caption{Confusion matrix for the evaluated classifier}
  \label{fig:conf}
    \end{center}
 \end{table}

\begin{figure*}
    \centering
    \includegraphics[width=\textwidth]{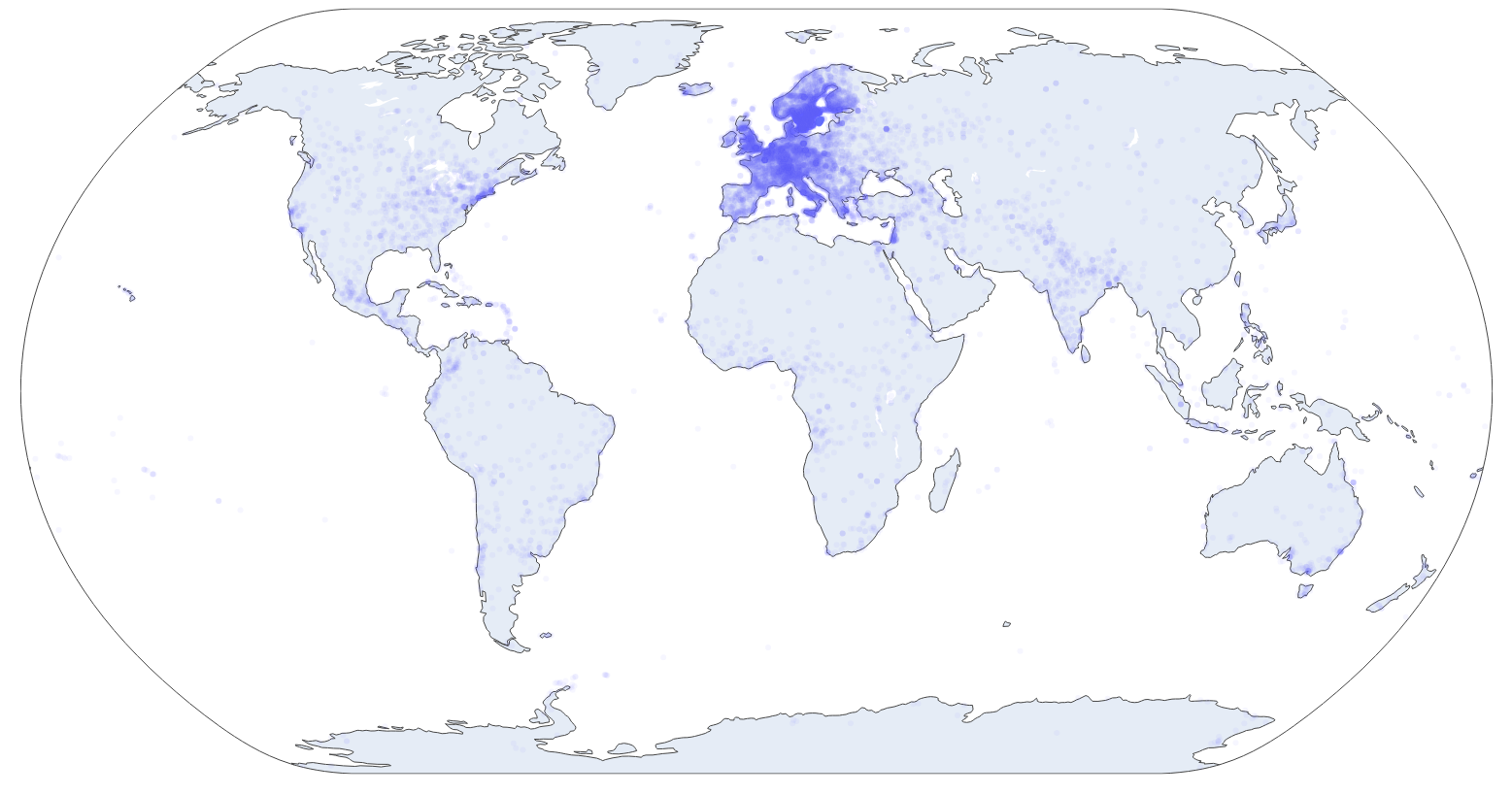}
    \caption{Visual representation of the encyclopedia's geographic locations}
    \label{fig:map}
\end{figure*}

Although not perfect, the classification reaches high results in all metrics. Nonetheless with a large corpus such as an encyclopedia, mislabeling of entries results in locations being ignored and incorrect entries being included. The precision score indicates that a high percentage of the identified locations are indeed valid geographic locations, while the recall score indicates that a high percentage of the actual geographic locations present in the encyclopedia were correctly identified and subsequently marked on the map, as shown in Figure~\ref{fig:map}. The F1 score also indicates that we strike a good balance between accurately identifying locations (precision) and not missing many valid locations (recall).

\begin{table}[tb]
\centering
\begin{tabular}{lr}
\hline
Accuracy  & 0.935  \\ 
Precision & 0.939  \\ 
Recall    & 0.930   \\ 
F1-score  & 0.935 \\ \hline
\end{tabular}
\caption{Performance metrics for the classifier.}
\label{metrics}
\end{table}

\subsection{Scraper}
There is a discrepancy between our scraping and the total number of entries in the encyclopedia. This is in part due to some subentries being under one entry, such as different members of one family all being in the surname. These are counted as one entry in our dataset. However, it may also be due to errors in the scraping process or to an incorrect optical character recognition.

\subsection{Error Analysis}
Our automatic annotation process removed the workload of manually annotating the encyclopedia with correct QIDs and coordinates. In certain cases, however, the annotator could miss linkages due to various factors. As an example, older spellings of location names that were valid prior to the Swedish spelling reform of 1906 were not always present as alternative name labels in Wikidata. This, in turn, led to an unsuccessful linkage between the encyclopedia entry and the corresponding Wikidata item that in reality should have been linked.

We discuss below three examples of errors, where the system did not correctly identify the QID.

\subsubsection{Aachen}
The encyclopedia has an \textit{Aachen} entry which reads:
\begin{quote}
    Aachen [ak-]. 1. Regeringsområde i preussiska
Rhenprovinsen, 4,155 kvkm. med 614,964 inv. (1900).

    2. (Lat. Aquisgranum, fr. Aix-la-Chapelle) Hufvudort
    i nyssnämnda område, vid den lilla ån Worm l. Wurm,
    nära gränsen till Holland och Belgien.

\end{quote}

translated as:

\begin{quote}
    Aachen [ak-]. 1. Government area in the Prussian Rhine Province, 4,155 sq km with 614,964 inhab. (1900).

    2. (Lat. Aquisgranum, fr. Aix-la-Chapelle) Capital
    in the area just mentioned, by the small river Worm l. Wurm,
    near the border with Holland and Belgium.
\end{quote}

In this case, the difficulty mainly lies in the fact that the encyclopedia entry for \emph{Aachen} has two distinct definitions: the first being the now defunct administrative district of the former German Empire (\verb=Q896929=) and the second being the city itself (\verb=Q1017=). 

Distinguishing between these two definitions and and linking them to the correct Wikidata entry automatically is not a trivial task. To classify these entries correctly, it must be ensured that the annotator is presented with sufficient context from the ambiguous entries in order to successfully choose the correct candidate for the entry. One method may be to employ a segmenter to construct subarticles and compare the different uses of a word. This is a potential idea for further work. 

\subsubsection{Arktonnesos}
Another entry of interest is \emph{Arktonnesos}, where the encyclopedia entry reads:

\begin{quote}
    Arktonnesos, det grekiska namnet på den i Marmarasjön utskjutande Artaki-halfön.
\end{quote}

translated as:

\begin{quote}
    Arktonnesos, the Greek name of the Artaki Peninsula which projects into the Sea of Marmara.
\end{quote}

The current name of this peninsula is \textit{Kapıdağ Peninsula}, which can be derived only through comparing the entry with external sources. The Wikidata item (\verb=Q3780284=) for Kapıdağ does not contain an alternative name that reads \emph{Arktonnesos} or any variations on that name. Consequently, this type of cases needs external reference sources to be mapped correctly. 

\subsubsection{Iowa}
The last example concerns the state of \emph{Iowa} entry:
\begin{quote}
Iowa, en af Nord-Amerikas förenta stater...
\end{quote}
translated as:
\begin{quote}
Iowa, one state of North America's United States...
\end{quote}
It showcases the importance of adequate context. 

For this entry, the linker chose  \verb=Q99670857= as QID, which corresponds to a fictional analog of this state. The entity description in Wikidata reads
\begin{quote}
    the federated state of Iowa in the USA as depicted in Star Trek
\end{quote}
where as the correct Wikidata entry description (\verb=Q1546=) reads:
\begin{quote}
    state of the United States of America
\end{quote}

This wrong link comes from the candidate ranking system for picking the QID. The cosine similarity between the SBERT embedding of the entry definition and the wikidata description was higher with the \textit{Star Trek} variant than the real-life counterpart.

We believe we could improve the results with a new ranking algorithm, with more constrains on the selection of candidates, or with a better encoding model. 
    
\section{Discussion and Conclusion}
As initial hypothesis for this work, we posed that the geographic locations mentioned in the encyclopedia would skew heavily towards Swedish places in particular, and European places in general. The results in Figure~\ref{fig:map} confirms this was correct. Furthermore, the encyclopedia features a large amount of geographic locations in Germany and the United Kingdom. These are countries where Sweden historically has had strong ties, especially Germany.

While the result may seem self-evident, it is nevertheless one from which further analysis could be built. For example, this study could, with little modification of the method, be extended to include later Swedish encyclopedias. Thus it would allow a comparative analysis on the choice of locations. After the Second World War for example, how would Germany and the United States evolve on the map? 

Some conclusions can be drawn in comparison to the geographic location linking and visualization of the French dictionary \textit{Le Petit Larousse Illustré} done in \citet{nugues2022connecting}. The \textit{Petit Larousse} was partially contemporary to \textit{Uggleupplagan}, and under the assumption that the selected locations are indicative of the respective countries' zeitgeist, some comparisons can be made. There is a higher density of locations in the United States, Japan and India, possibly due to the more global nature of France's transnational relations and trade at the time, as compared to Sweden's. The French dictionary also includes many entries in the former French colonies of French North and West Africa. 

\section{Acknowledgments}
This work was partially supported by \textit{Vetenskaprådet}, the Swedish Research Council, registration number 2021-04533.

\nocite{*}
\section{Bibliographical References}\label{sec:reference}

\bibliographystyle{lrec-coling2024-natbib}
\bibliography{bibliography}

\end{document}